\documentclass[conference, letterpaper]{IEEEtran}
\IEEEoverridecommandlockouts
\usepackage{cite}
\usepackage{amsmath,amssymb,amsfonts}
\usepackage{graphicx}
\usepackage{textcomp}
\usepackage{xcolor}
\def\BibTeX{{\rm B\kern-.05em{\sc i\kern-.025em b}\kern-.08em
    T\kern-.1667em\lower.7ex\hbox{E}\kern-.125emX}}
\begin{document}

\title{Leveraging a Foundation Model for the EEG-Based Diagnosis of Alzheimer's Disease \\
\thanks{ EEG device and data collection was supported by HippoScreen Neurotech Corp. 30 channels were indexed according to the international 10–20 system as follows:
1: T5, 2: T3, 3: F7, 4: O1, 5: CP3, 6: FC3, 7: Fp1, 8: FCz, 9: CPz, 10: Oz,
11: F4, 12: C4, 13: P4, 14: FT8, 15: TP8, 16: TP7, 17: FT7, 18: P3,
19: C3, 20: F3, 21: Fz, 22: Cz, 23: Pz, 24: Fp2, 25: FC4,
26: CP4, 27: O2, 28: F8, 29: T4, 30: T6 }
}

\author{\IEEEauthorblockN{Maggie Lin}
\IEEEauthorblockA{\textit{Department of Bioengineering}\\
\textit{University of California, San Diego}\\
\textit{La Jolla, CA 92037, USA}\\
mal085@ucsd.edu}
\and
\IEEEauthorblockN{Chung-Lin Hou}
\IEEEauthorblockA{\textit{Research and Development}\\
\textit{HippoScreen Neurotech Corp.}\\
\textit{Taipei, Taiwan}\\
jonathan\_hou@hipposcreen-nc.com}
\and
\IEEEauthorblockN{Tzyy-Ping Jung}
\IEEEauthorblockA{\textit{Swartz Center for Computational Neuroscience}\\
\textit{University of California, San Diego}\\
\textit{La Jolla, CA 92037, USA}\\
tpjung@ucsd.edu}
}

\maketitle

\begin{abstract}
Biological heterogeneity in Alzheimer’s Disease (AD) poses a critical diagnostic challenge, particularly for traditional linear methods that fail to capture non-linear neural dynamics. To address this, we propose a diagnostic framework utilizing the Large Brain Model (LaBraM), pretrained on over 2,500 hours of EEG data. By integrating these high-dimensional latent embeddings with a non-linear Random Forest classifier, our approach effectively isolates robust disease markers. Under a rigorous subject-independent 5-fold cross-validation protocol, the method achieves an ROC-AUC of 89.36\% ± 3.49\%, PR AUC of 81.45\% ± 4.43\%, and Balanced Accuracy of 82.44\% ± 4.34\% in distinguishing dementia patients from healthy controls. Notably, this performance uses only 8-second EEG segments, surpassing traditional spectral baselines, including band-power and parameterized oscillatory features (FOOOF). Post-hoc occlusion analysis confirms the model captures clinically validated biomarkers, specifically occipital-frontal Alpha and Theta rhythm degradation. Additional neurophysiological alignment analysis demonstrated that higher LaBraM-predicted dementia probability significantly correlated with worse cognitive performance, greater clinical severity, increased theta and alpha relative power, and higher aperiodic exponent. These findings demonstrate that deep latent representations extract clinically relevant signatures from noisy signals, enabling precise, rapid, and data-efficient diagnosis.
\end{abstract}


\section{Introduction}

Alzheimer’s Disease (AD) is a progressive neurodegenerative disorder characterized by significant biological heterogeneity, making early and accurate diagnosis a critical challenge in modern neurology. While Electroencephalography (EEG) offers a non-invasive and cost-effective window into neural activity, detecting the subtle, prodromal signatures of dementia remains difficult due to the low signal-to-noise ratio and the complex, non-stationary nature of brain dynamics. Traditional analysis has largely relied on hand-crafted spectral features, such as Relative Band-Power or the Alpha-to-Theta ratio\cite{lee2025exploring}. While clinically interpretable, these linear methods inherently oversimplify the rich spatiotemporal structure of EEG signals. Crucially, preliminary analysis in this study revealed that these conventional metrics yielded no statistically significant differences ($p > 0.05$) between early-stage AD and healthy controls, highlighting a discriminative bottleneck in which traditional manual feature engineering fails to capture subtle pathological shifts. Consequently, a diagnostic framework capable of decoding high-dimensional interactions is required.

Deep Learning (DL) facilitates data-driven representation learning but typically requires massive labeled datasets to generalize\cite{xia2023novel}. Large Brain Models (LaBraM) address this scarcity by leveraging self-supervised pre-training on over 2,500 hours of unlabeled EEG data\cite{jiang2024labram},\cite{wang2025theoryapplicationfinetuninglarge}. These foundation models learn robust, context-aware neural representations, offering an opportunity to extract generalized spatiotemporal features inaccessible to conventional spectral decomposition.  

In this study, we propose a geometric framework that integrates the pre-trained power of LaBraM with non-linear classification for robust AD detection. Unlike previous approaches that rely on black-box end-to-end training, we explicitly hypothesize that AD-related biomarkers reside within the high-dimensional LaBraM latent space and are better captured by nonlinear decision boundaries. To validate this, we implement a LaBraM-Random Forest pipeline, benchmarking it against two rigorous spectral baselines: traditional Band-Power and the modern Fitting Oscillations \& One-Over-F (FOOOF) parameterization\cite{lee2025exploring},\cite{gerster2022separating}. 
\begin{figure*}[t]
    \centering
    \includegraphics[
        width=1\textwidth,
        trim={0 0.90in 0 0.90in},
        clip
    ]{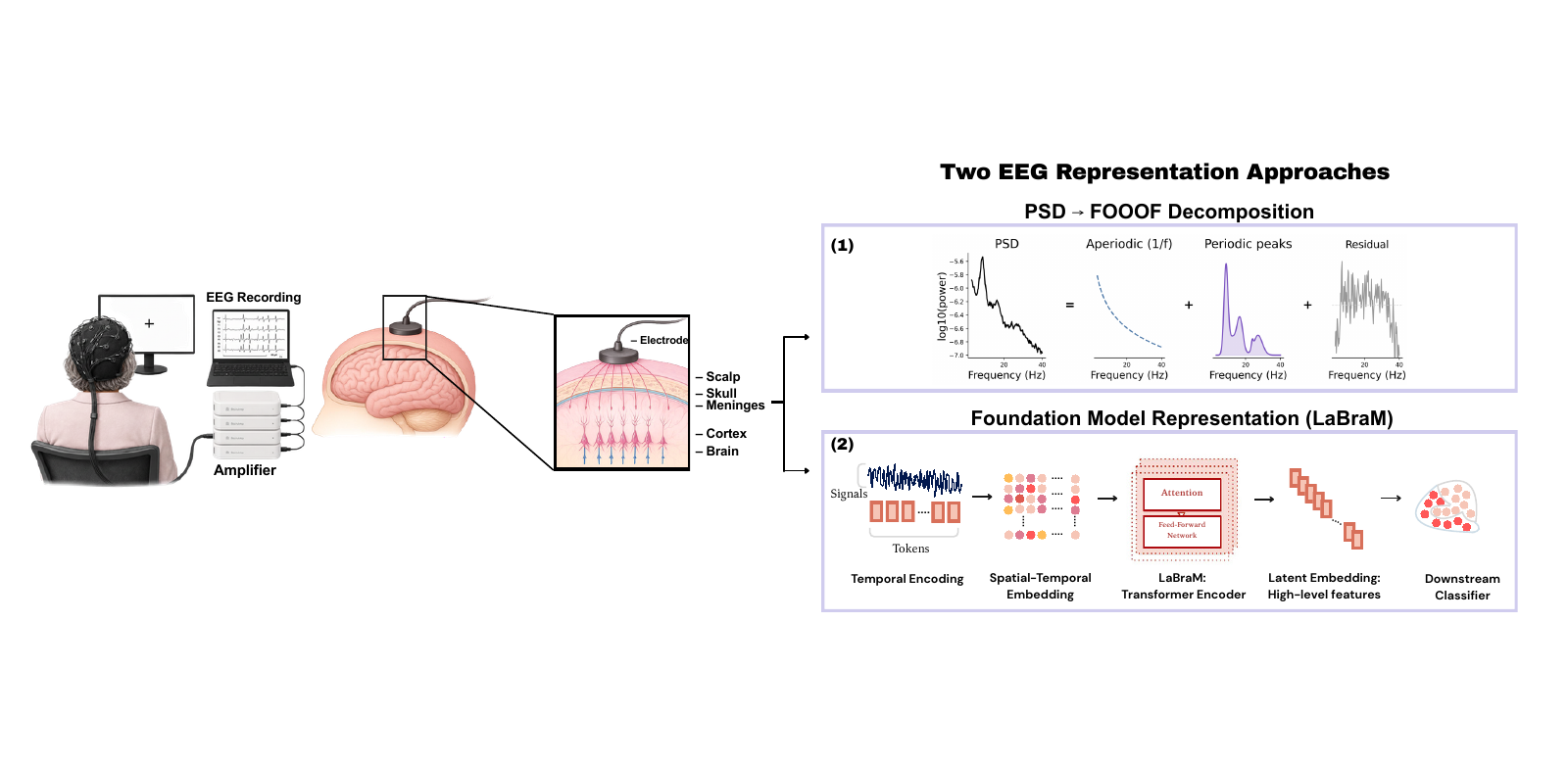}
    \caption{Overview of the proposed EEG-based diagnostic framework. Scalp EEG signals are recorded and analyzed using two complementary representation approaches: \textbf{(1)} interpretable spectral biomarkers and \textbf{(2)} foundation-model embeddings. Power spectral density (PSD) features are decomposed by FOOOF into aperiodic and periodic components, while EEG epochs are tokenized and embedded by the pretrained LaBraM Transformer encoder. The extracted latent representations are then used for downstream dementia classification.}
    \label{fig:pipeline}
\end{figure*}
\begin{figure*}[t] 
    \centering {\includegraphics[width=1\linewidth]{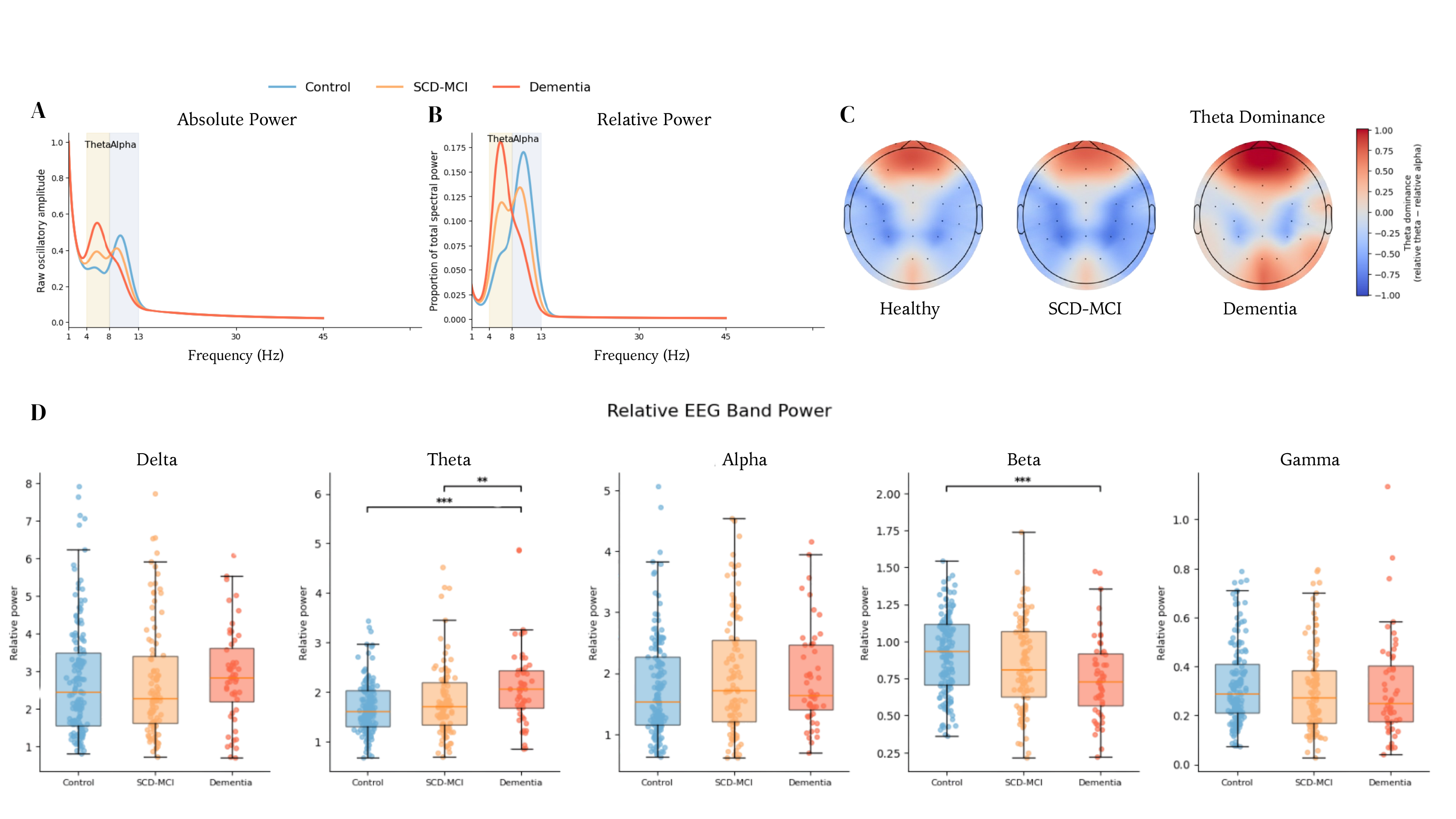}}
    \caption{Power spectral density analysis across diagnostic groups. PSD was estimated from cleaned EEG recordings and summarized as absolute and relative power. \textbf{(a)} Absolute power spectra show increased theta-band activity as disease severity progresses. \textbf{(b)} Relative power spectra highlight a shift from alpha-dominant activity in Healthy subjects toward theta-dominant activity in Dementia patients. \textbf{(c)} Theta-dominance topographic maps show progressively stronger theta dominance from Healthy to SCD-MCI and Dementia groups, with red regions indicating greater relative theta than alpha activity. \textbf{(d)} Relative band-power comparisons show significant group differences in theta and beta bands, with increased theta power and reduced beta power in Dementia patients.}
    \label{fig:psd}
\end{figure*}
\section{Materials and Methods}

\subsection{Proposed Methodological Framework}
We benchmarked our proposed framework against two distinct spectral paradigms: (1) traditional Band-Power features and (2) the FOOOF algorithm, to evaluate the performance of (3) pretrained LaBraM embeddings.

\subsubsection{Large Brain Model (LaBraM)}
Fig. \ref{fig:pipeline} (2) illustrates the LaBraM framework. We first segment and tokenize raw continuous EEG signals into discrete patches. These tokens are projected into a high-dimensional latent space using pre-trained embeddings, which are enriched with temporal and spatial positional encodings to preserve the chronological order and topological geometry of the electrode array\cite{jiang2024labram}.

The resulting embeddings serve as input to a deep transformer encoder architecture. Within each layer, Multi-Head Self-Attention (MHSA) mechanisms and Feed-Forward Networks (FFN) are applied iteratively to capture long-range dependencies and complex non-linear interactions\cite{cordonnier2021multiheadattentioncollaborateinstead},\cite{raffel2016feedforwardnetworksattentionsolve}. Following the forward pass through the pre-trained layers, the model outputs high-level, context-aware feature vectors that encode the sophisticated latent dynamics of the brain signals.

\subsubsection{Random Forest}
Random Forest was used as a nonlinear ensemble classifier to capture complex interactions among LaBraM embedding features. The model builds multiple decision trees from bootstrapped training samples and random feature subsets, allowing nonlinear partitioning of the latent EEG space while reducing overfitting through ensemble averaging. This is well-suited for high-dimensional LaBraM embeddings, where
dementia-related information may be distributed across multiple latent dimensions. Hyperparameters, including the number of trees, maximum depth, minimum samples per leaf, and maximum feature subset, were tuned using validation data. Balanced class weights were applied to account for class imbalance, and the final prediction threshold was selected to maximize validation balanced accuracy.

\subsubsection{Dimensionality Reduction via Principal Component Analysis (PCA)}
The proximity of our sample size ($N=206$) to the embedding dimension ($d=200$) results in a sparse feature space, forcing classifiers to fit stochastic noise rather than true disease signals. To resolve this, we projected the embeddings into a compact, uncorrelated subspace using PCA with whitening. By retaining 95\% of the explained variance, we compressed the feature vector from 200 to 44 dimensions. This 78\% reduction filters noise while preserving robust pathological signatures. Notably, the whitening step isotropicizes the feature covariance. While this optimizes the decision boundary for the Support Vector Machine, we ensured that the transformation did not inadvertently remove variance required by the Random Forest for optimal recursive partitioning.

\subsubsection{Support Vector Machine (SVM) with a Radial Basis Function (RBF)}
To mitigate class imbalance, the SVM objective function was weighted inversely to class frequency, ensuring the model penalized misclassifications of the minority Dementia group more than those of the majority Control group\cite{scikit-learn}.

For model generalization, the hyperparameters, specifically the regularization strength $C$ and the RBF kernel coefficient $\gamma$, were optimized on an independent validation set. The optimal parameter combination was selected based on the maximum balanced accuracy achieved on the validation split. Subsequently, the final model was evaluated on a strictly isolated test set to report unbiased performance metrics.

For the classification task, we employed a SVM with RBF kernel, defined as:

\begin{equation}
    K(\mathbf{x}_i, \mathbf{x}_j) = \exp\left(-\gamma \|\mathbf{x}_i - \mathbf{x}_j\|^2\right),
    \label{eq:rbf_kernel}
\end{equation}
where $\mathbf{x}_i$ and $\mathbf{x}_j$ are feature vectors in the input space, and $\gamma$ is a hyperparameter controlling the influence of individual training samples.

\subsection{Baseline Feature Extraction for Comparison}
\subsubsection{Band-Power}
As the traditional clinical baseline, we computed the standard Power Spectral Density (PSD) across five canonical physiological ranges: Delta (0.5–4 Hz), Theta (4–8 Hz), Alpha (8–13 Hz), Beta (13–30 Hz), and Gamma (30–45 Hz). This approach is designed to capture the significant increase in low-frequency power (Delta and Theta) and the concomitant decrease in higher-frequency oscillatory activity (Alpha, Beta, and Gamma). 

\subsubsection{Spectral Parameterization (FOOOF)}
To overcome the limitations of fixed-band analysis, we also employed the FOOOF algorithm to systematically decompose the neural power spectra into periodic and aperiodic components. We extracted the aperiodic offset and exponent, alongside the center frequency, peak power, and bandwidth of the periodic oscillatory peaks. The model fitting was restricted to the 3–40 Hz range and optimized using the Bayesian Information Criterion (BIC) to ensure a robust representation while preventing overfitting. By separating the 1/f-like background activity from true physiological oscillations, FOOOF provides a more granular parameterization of neural dynamics than band power methods\cite{gerster2022separating},\cite{Wilson2024.08.01.606216}.

Both baseline feature sets were classified using Random Forest (RF) and radial basis function Support Vector Machine (RBF-SVM) classifiers. These baselines were subjected to the same subject-independent 5-fold cross-validation framework as the proposed LaBraM-based model, with all results reported as mean $\pm$ standard deviation to ensure a fair comparison.

\subsection{Experimental Setup and Data Preprocessing}

\subsubsection{Data Acquisition and Partitioning} 30-channel EEG data (120 seconds resting-state, eyes-open) were collected between 2021–2022 across four medical centers in Taiwan. The cohort comprised 308 subjects divided into Control (Healthy and Subjective Cognitive Decline), Mild Cognitive Impairment (MCI) and Dementia groups, with clinical diagnoses serving as ground truth. Diagnostic characterization was supported by standard clinical assessments, including the Mini-Mental State Examination (MMSE), Montreal Cognitive Assessment (MoCA), Clinical Dementia Rating (CDR), Geriatric Depression Scale (GDS), Instrumental Activities of Daily Living (IADL), and Activities of Daily Living (ADL). For downstream binary classification,  MCI patients ($N$ = 102) were excluded, resulting in a Control-Dementia cohort of 206 subjects. The dataset was partitioned into training ($N=105$), validation ($N=38$), and independent testing ($N=63$) sets. The fixed split was used for hyperparameter tuning, whereas 5-fold cross-validation on the combined subject-level pooled embedding dataset was used for final evaluation.

\subsubsection{Data Preprocessing}
Raw EEG signals recorded at 500 Hz were bandpass filtered (0.1–50 Hz) and downsampled to 200 Hz. The subsequent pipeline addressed artifacts and scaling: (1) noisy frontopolar channels were reconstructed using spherical spline interpolation\cite{ferree2006spherical}; (2) Independent Component Analysis (ICA) removed ocular, myogenic, and cardiac artifacts\cite{makeig1995independent}; (3) signals were re-referenced to the Common Average Reference (CAR)\cite{ludwig2009using}; and (4) channel-wise Z-score normalization was applied prior to feature extraction\cite{apicella2023effects}.

\subsubsection{Spectral Biomarker Analysis Using PSD and FOOOF} Before extracting foundation-model representations, conventional spectral biomarkers were analyzed across clinical stages through the following steps: (1) power spectral density (PSD) was estimated using Welch’s method \cite{welch1967use}; (2) absolute and relative band-power features were extracted across canonical EEG bands, with relative power normalized by total 1–45 Hz power; and (3) theta-dominance measures were then derived to quantify EEG slowing; and (4) group-level statistical testing was performed on relative band-power features to identify diagnostic differences. To further determine whether these spectral differences reflected oscillatory activity or broadband background shifts, FOOOF was applied to decompose the power spectrum into periodic and aperiodic components. MCI groups were retained only for exploratory spectral-stage analysis and were excluded from the downstream binary classification task. 

\subsubsection{Data Augmentation and Classification} Clean EEG signals were processed in the following sequence: (a) segmented the cleaned data into 8-second epochs; (b) applied a 75\% overlap (2-second stride) exclusively to the Dementia group, expanding the number of epochs, while using non-overlapping windows for the Control group \cite{wang2025theoryapplicationfinetuninglarge},\cite{martins2023data}; (c) extracted LaBraM embeddings for each epoch; (d) pooled embeddings at the subject level; and (e) performed classification on subject-level pooled LaBraM embeddings using Random Forest, Logistic Regression, and SVM classifiers. Because classification was performed on subject-level pooled embeddings, overlapping segments were not treated as independent samples in the final classifier. 

\subsubsection{Statistical Analysis} Model performance was evaluated using 5-fold cross-validation on the subject-level pooled embedding dataset, such that each sample corresponded to one subject. The confusion matrix was generated by aggregating predictions from the held-out subjects across all five cross-validation folds. Each iteration utilized 4 folds for training and 1 fold for testing, and performance metrics are reported as the mean ± standard deviation across the 5 folds. Statistical significance of the proposed framework against baselines was assessed using one-tailed paired t-tests on ROC-AUC scores.

\subsubsection{Clinical Interpretability} To demystify the "black box" nature of the LaBraM transformer, three post-hoc interpretability analyses were performed: (1) Band-stop filters were applied to five standard EEG bands (Delta, Theta, Alpha, Beta, and Gamma) before feature extraction. Importance was quantified by the ROC AUC drop relative to baseline performance. (2) A-single-channel occlusion test zeroed out each of the 30 channels individually. The resulting AUC declines identified the top 10 most influential sensors, pinpointing critical anatomical regions for classification. (3) clinical and neurophysiological alignment was assessed by correlating LaBraM-predicted dementia probability with clinical assessment scores and EEG-derived physiological features, including relative band power and FOOOF-derived aperiodic components, using Spearman correlation with false discovery rate correction.

\section{Experimental Results}

 \begin{figure}[t!]  
    \centering
    \includegraphics[width=\columnwidth]{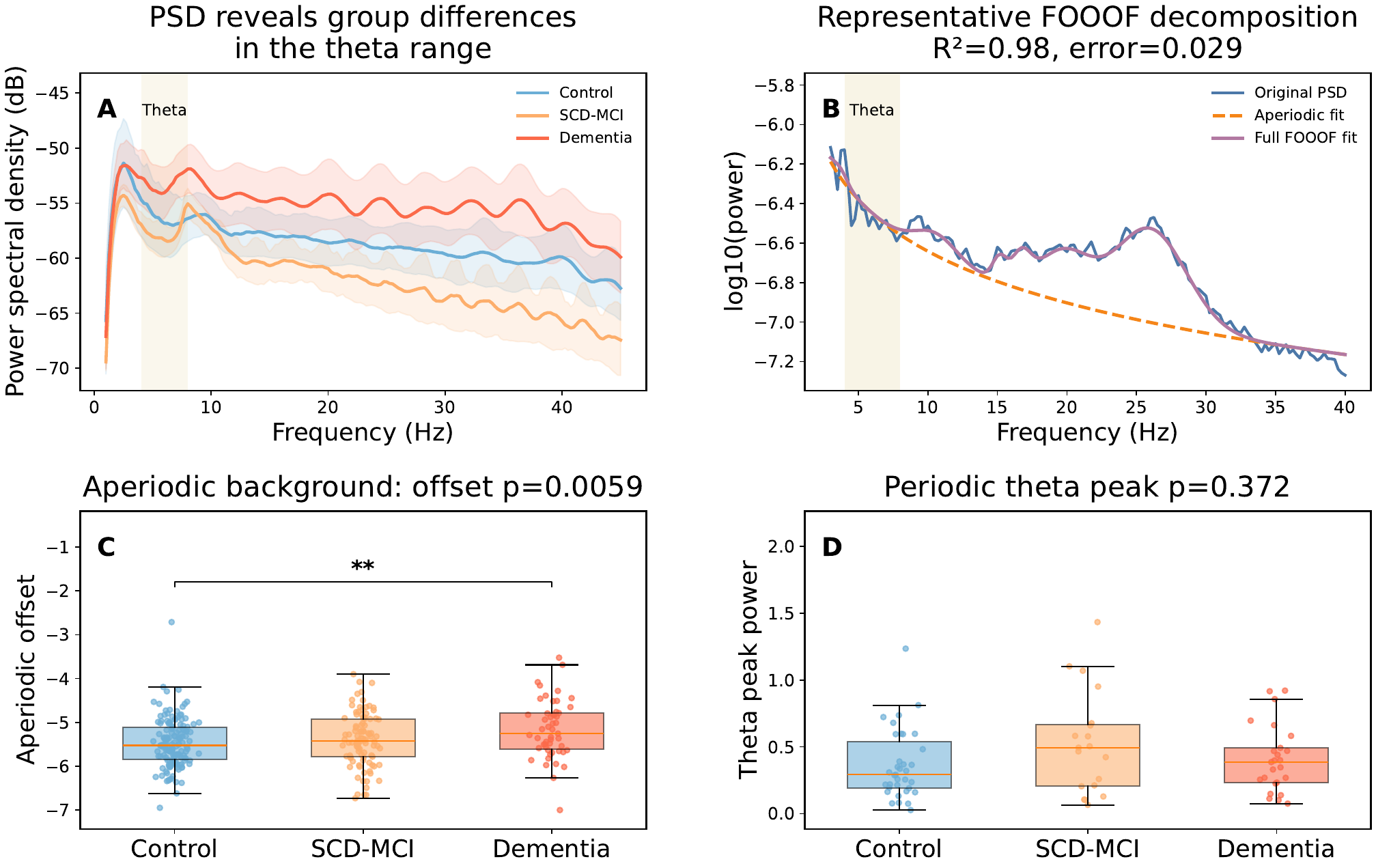} \caption{Decomposing theta-band PSD differences into background and peak components. \textbf{(a)} PSD analysis shows group differences in the theta range across diagnostic groups. \textbf{(b)} A representative FOOOF fit separates the power spectrum into aperiodic background and periodic peak components. \textbf{(c)} Aperiodic offset showed a significant group difference (p = 0.0059), whereas \textbf{(d)} periodic theta peak power did not differ significantly across groups (p = 0.372). These results suggest that the theta-band PSD difference is mainly driven by aperiodic background changes rather than isolated theta oscillatory peak power.}
    \label{fig:fooof}
\end{figure}

\subsubsection{PSD and FOOOF biomarkers}

Fig. \ref{fig:psd} shows disease-related spectra by diagnostic groups. Theta-dominance maps and relative band-power analysis demonstrated increased theta activity in Dementia patients, consistent with EEG slowing, while beta power was significantly reduced. However, FOOOF decomposition in Fig. \ref{fig:fooof}(c-d)  further indicated that the theta-band alteration was primarily driven by aperiodic offset changes rather than periodic theta peak power, suggesting that dementia-related EEG slowing reflects broader background spectral shifts beyond conventional band-power differences. 

\begin{table*}[htbp]
\centering
\caption{Classification Performance Comparison (Mean $\pm$ Standard Deviation)}
\label{tab:results}{
\begin{tabular}{llccccc}
\hline
\multicolumn{7}{c}{\textbf{Model Performance}}\\
\hline

\textbf{\textit{Feature}}& 
\textbf{\textit{Classifier}} & 
\textbf{\textit{Balanced}} & 
\textbf{\textit{ROC AUC (\%)}} & 
\textbf{\textit{PR AUC (\%)}} & 
\textbf{\textit{Specificity (\%)}} & 
\textbf{\textit{Sensitivity (\%)}} \\

\textbf{\textit{Extraction}}& 
& 
\textbf{\textit{Accuracy (\%)}} & 
& 
& 
\textbf{\textit{(Control)}} & 
\textbf{\textit{(Dementia)}} \\
\hline

\multicolumn{7}{l}{\textbf{Fixed Split (Train/Valid/Test)}}\\
\hline

LaBraM & Logistic Regression & 58.64 & 58.78 & 32.49 & 29.79 & 87.50 \\
LaBraM & SVM (Linear) & 58.64 & 50.27 & 24.71 & 29.79 & 87.50 \\
LaBraM & SVM (RBF) & 80.12 & 84.44 & 80.10 & 91.49 & 68.75 \\
LaBraM& PCA + SVM (RBF)& 81.18& 85.90& 79.60& 93.62& 68.75\\
 \textbf{LaBraM}& \textbf{Random Forest}$^{\mathrm{1}}$& \textbf{80.12}$^{\mathrm{**}}$& \textbf{93.09}$^{\mathrm{**}}$& \textbf{79.47}$^{\mathrm{**}}$& \textbf{91.49}$^{\mathrm{**}}$&\textbf{68.75}$^{\mathrm{**}}$\\
\hline

\multicolumn{7}{l}{\textbf{Subject-independent 5-fold cross-validation}}\\
\hline

FOOOF & SVM (RBF) & 54.73 $\pm$ 9.13 & 59.87 $\pm$ 7.79 & 40.18 $\pm$ 6.65 & 55.82 $\pm$ 5.85 & 53.64 $\pm$ 21.20 \\
Band Power & Random Forest & 60.07 $\pm$ 8.52 & 65.09 $\pm$ 11.28 & 47.42 $\pm$ 11.76 & 62.32 $\pm$ 3.43 & 57.82 $\pm$ 16.11 \\
 LaBraM & Logistic Regression & 48.06 $\pm$ 9.96 & 48.59 $\pm$ 5.26 & 28.77 $\pm$ 3.16 & 27.40 $\pm$ 12.46 &68.73 $\pm$ 32.19 \\
 LaBraM & SVM (Linear) & 54.55 $\pm$ 1.76 & 47.25 $\pm$ 6.74 & 27.97 $\pm$ 3.79 & 47.83 $\pm$ 31.74 &61.27 $\pm$ 29.86 \\
LaBraM & SVM (RBF) & 72.17 $\pm$ 9.63 & 82.23 $\pm$ 10.50 & 76.38 $\pm$ 12.31 & 88.34 $\pm$ 10.10 & 56.00 $\pm$ 24.87 \\
LaBraM & PCA + SVM (RBF) & 76.69 $\pm$ 7.81 & 81.69 $\pm$ 5.44 & 74.00 $\pm$ 8.56 & 83.20 $\pm$ 12.78 & 70.18 $\pm$ 21.31 \\
\textbf{LaBraM} & \textbf{Random Forest}$^{\mathrm{1}}$ & \textbf{82.44 $\pm$ 4.34}$^{\mathrm{**}}$& \textbf{89.36 ± 3.49}$^{\mathrm{**}}$& \textbf{81.45 $\pm$ 4.43}$^{\mathrm{**}}$& \textbf{85.78 $\pm$ 7.77}$^{\mathrm{**}}$& \textbf{79.09 $\pm$ 8.96}$^{\mathrm{**}}$\\
\hline

\multicolumn{7}{l}{$^{\mathrm{1}}$Best Performance.}\\
\multicolumn{7}{l}{$^{\mathrm{**}}$Statistically significant improvement ($p < 0.01$) over baseline using a paired t-test.}\\
\hline

\end{tabular}
}
\end{table*}

 \begin{figure}[t!]  
    \centering
    \includegraphics[width=\linewidth]{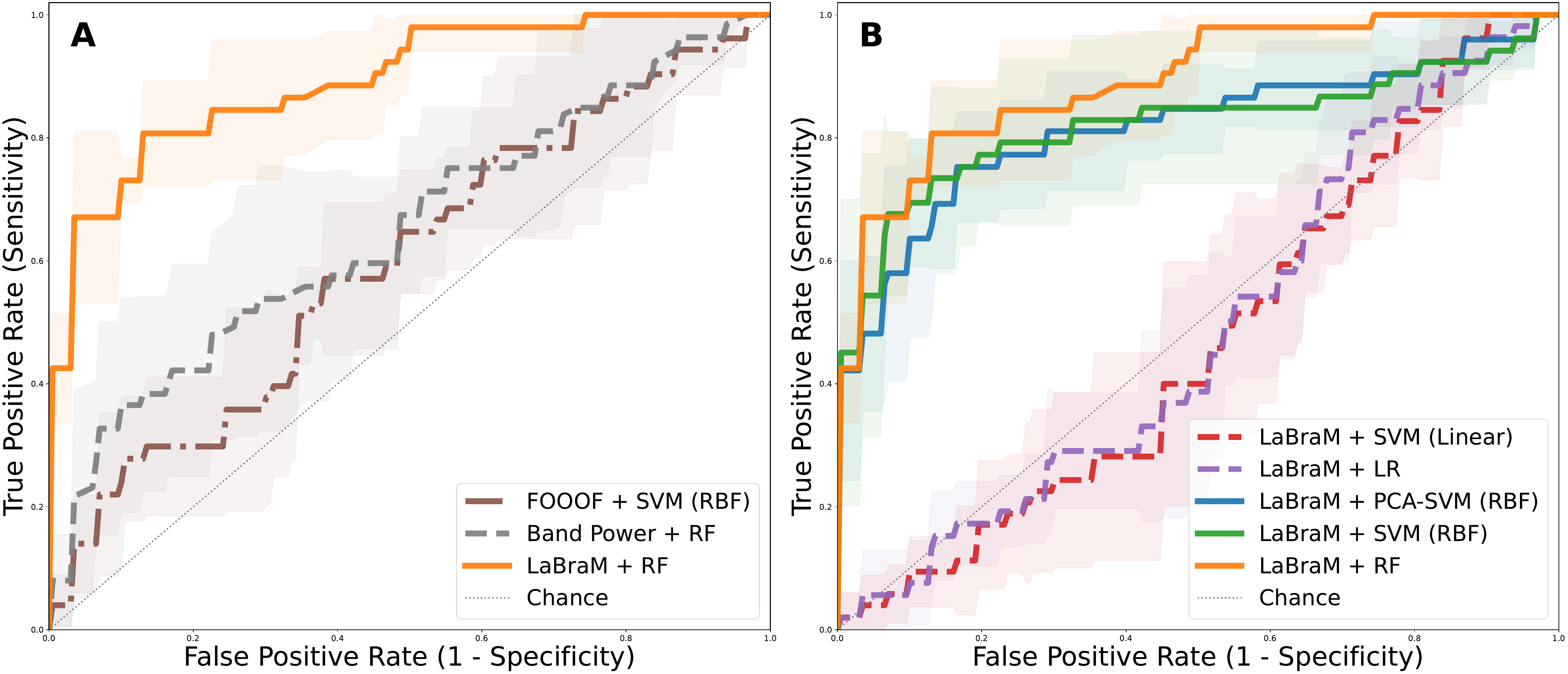}
    \caption{ \textbf{(a)} Comparison of feature extraction methods, demonstrating that LaBraM features (AUC = 89.4\% ± 3.5\%) outperform Band-Power (AUC = 65.1\% ± 11.3\%) and FOOOF (AUC= 59.9\% ± 7.8\%).  \textbf{(b)} Performance evaluation of different classifiers, where the proposed LaBraM with Random Forest achieves the highest AUC of 89.4\% ± 3.5\%, surpassing SVM-RBF (AUC = 82.2\% ± 10.5\% ), PCA-SVM-RBF (AUC = 81.7\% ± 5.4\% ), Logistic Regression (AUC = 48.6\% ± 5.3\%), and SVM-Linear (AUC = 47.2\% ± 6.7\%).}
    \label{fig:roc_curves}
\end{figure}
 \begin{figure}[t!]  
    \centering
    \includegraphics[width=0.30\textwidth]{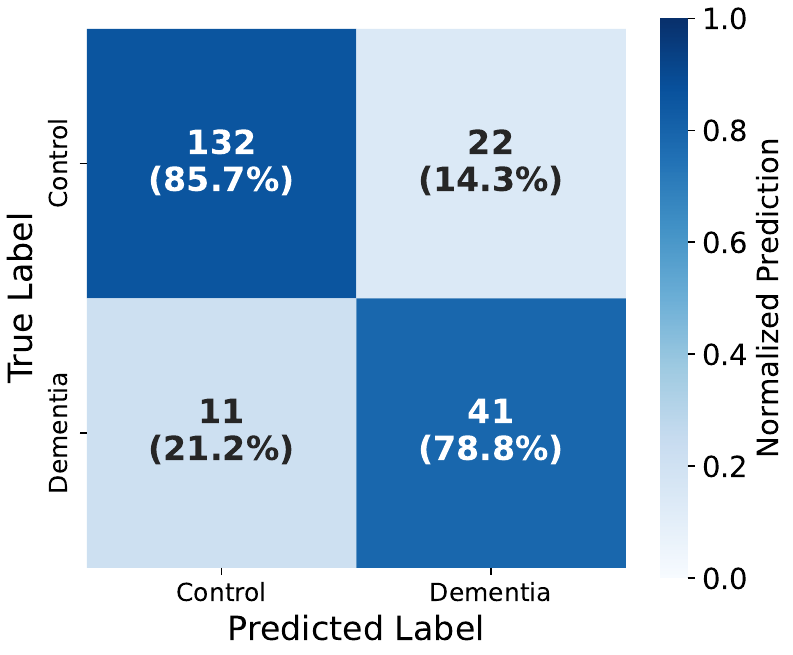} 
    \caption{Aggregated Confusion matrix across 5-fold cross-validation: The model achieves a Specificity of 85.7\% for the Healthy control group and a Sensitivity of 78.8\% for the Dementia group.}
    \label{fig:confusion_matrix}
\end{figure}

\begin{figure}[htbp]
\centerline{\includegraphics[width=1\linewidth]{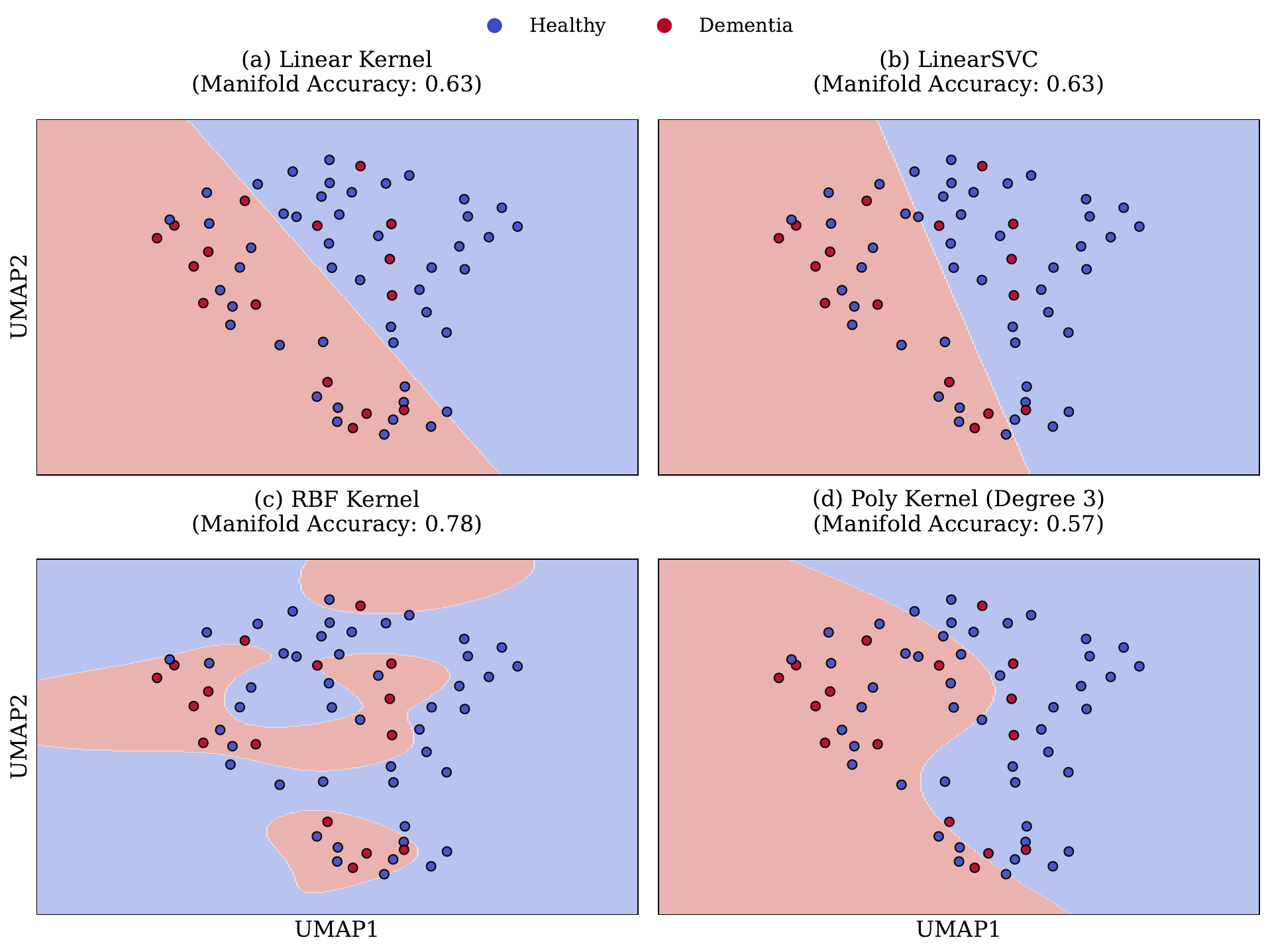}}
\caption{The UMAP parameters were set to n neighbors = 10 and min dist = 0.1 to strictly preserve the local neighborhood relationships while clearly separating the clusters. Within this reduced 2D space, we retrained and visualized the decision boundaries of four comparative kernels (Linear, LinearSVC, Poly, and RBF) to qualitatively assess the linear separability of the dementia biomarkers.  \textbf{(a) }Linear Kernel and\textbf{ (b) }LinearSVC show limited separability (Manifold Accuracy = 0.63). \textbf{(c) }The RBF Kernel demonstrates superior performance (Accuracy = 0.78), effectively separating the classes in the low-dimensional space. \textbf{(d) }Polynomial Kernel (Degree 3) yields the lowest accuracy of 0.57. }
\label{SVM_UMAP}
\end{figure}
\subsubsection{Classifier Performance}

Table \ref{tab:results} and Fig. \ref{fig:roc_curves} summarize the 5-fold cross-validation for the classification performance. The proposed LaBraM-RF framework achieved a superior ROC-AUC of 89.4\% ± 3.5\%, surpassing the PCA-SVM-RBF (AUC = 81.7\% ± 5.4\%) and standard SVM-RBF (82.2\% ± 10.5\%). Conversely, linear classifiers (Logistic Regression and Linear SVM) performed near chance level of AUC =  48.6\% ± 5.3\% and AUC = 47.3\% ± 6.7\% respectively, validating the hypothesis that the pathological features reside on a non-linear space. As detailed in the aggregated confusion matrix across the 5-fold validation in Fig. \ref{fig:confusion_matrix}, the model demonstrates exceptional clinical reliability, achieving a balanced profile with a Sensitivity of 78.8\% for correctly identifying Dementia patients and a Specificity of 85.7\% for accurately recognizing Healthy controls. 
\begin{figure*}[t] 
    \centering {\includegraphics[width=1\linewidth]{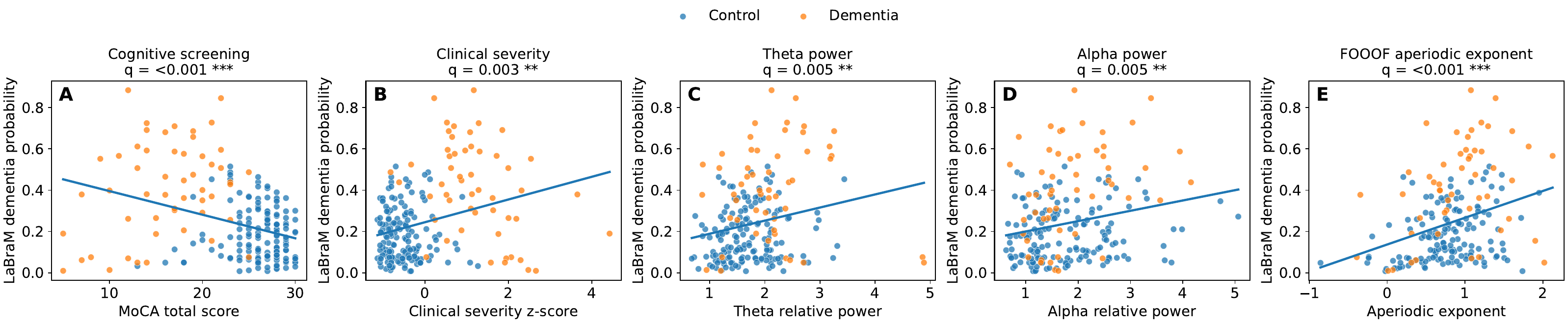}}
    \caption{Neurophysiological alignment of LaBraM model output. Higher LaBraM-predicted dementia probability was associated with \textbf{(a)} lower MoCA scores, reflecting worse cognitive performance; \textbf{(b)} greater clinical severity; \textbf{(c--d)} increased theta and alpha relative power; and \textbf{(e)} higher aperiodic exponent after false discovery rate (FDR) correction. These findings suggest that LaBraM embeddings capture clinically meaningful EEG representations aligned with cognitive impairment and interpretable neurophysiological markers.}
    \label{fig:ai_neuro}
\end{figure*}

\subsubsection{Statistical Benchmarking}
A one-tailed paired $t$-test ($N=5$) on cross-validation AUC scores confirmed that the proposed framework yields statistically significant improvements over all baselines. Specifically, Fig. \ref{fig:roc_curves} highlights the statistical advantage over the top spectral power baseline (Band Power with Random Forest, $p = 0.004$) and the parameterized spectral baseline (FOOOF with SVM-RBF, $p = 0.009$).

\subsubsection{Linear versus Non-Linear Classifier}
Comparison against the linear ablation model (SVM-Linear) revealed a highly significant performance gap ($p < 0.01$), confirming that linear decision boundaries are insufficient to resolve the pathological heterogeneity of Dementia. This is corroborated by the UMAP visualization in Fig. \ref{SVM_UMAP}, where the healthy and dementia clusters exhibit significant non-linear overlap. Consequently, linear classifiers are constrained to a manifold accuracy of 54\%, whereas the RBF kernel successfully contours the complex decision surface, achieving a manifold accuracy of 79\%. 

\subsubsection{Post-hoc analysis of LaBraM and neurophysiology correlation}
To evaluate the clinical interpretability of LaBraM, we examined whether model-predicted dementia probability was associated with clinical impairment and EEG-derived physiological features. As shown in Fig. \ref{fig:ai_neuro}, higher LaBraM-predicted dementia probability was significantly associated with lower MoCA scores, reflecting worse cognitive performance, greater clinical severity, increased theta-alpha relative power, and higher aperiodic exponent. These results suggest that LaBraM embeddings capture clinically meaningful EEG representations aligned with cognitive impairment and interpretable neurophysiological markers. 

\subsubsection{Post-hoc analysis of Influential Brain Regions and Frequency Bands}
\begin{figure}[htbp]
\centerline{\includegraphics[width=1\linewidth]{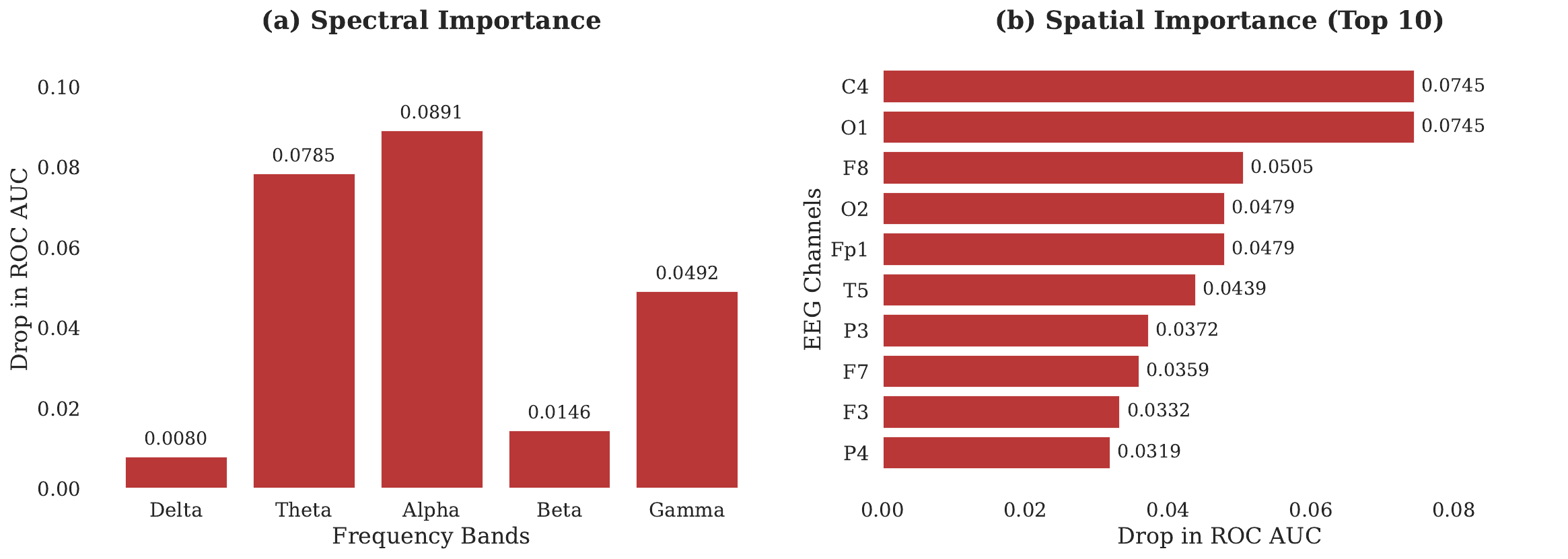}}
\caption{Feature importance analysis of the proposed LaBraM-RF pipeline.\textbf{ (a) }Spectral Importance: Results of frequency band occlusion demonstrating the model's primary reliance on Alpha (8–13 Hz) and Theta (4–8 Hz) rhythms. Filtering these bands resulted in ROC AUC drops of 8.91\% and 7.85\%, respectively. \textbf{(b)} Spatial Importance: The top 10 most influential EEG channels were identified through single-channel occlusion. C4 and O1 emerge as the most critical spatial predictors; removing either channel leads to a 7.45\% decline in ROC AUC.}
\label{importance}
\end{figure}

\begin{figure}[htbp]
\centerline{\includegraphics[width=0.5\linewidth]{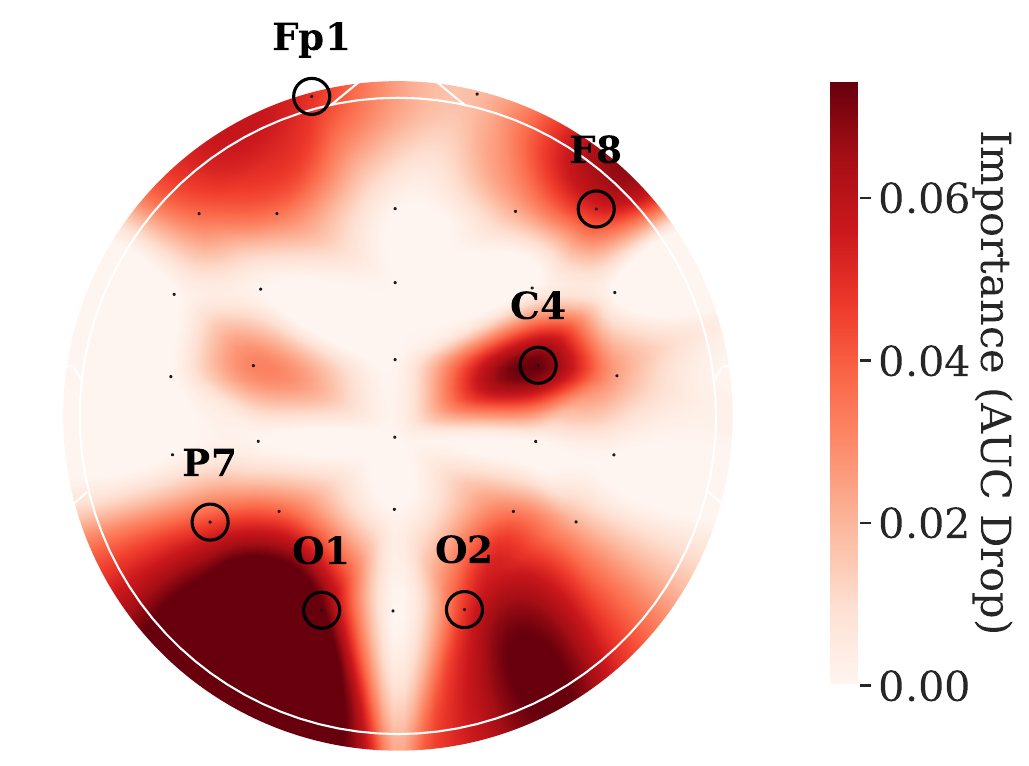}}
\caption{Topographic distribution of feature importance for AD diagnosis. The heatmap illustrates the drop in AUC when individual EEG channels are occluded. Warmer colors indicate higher importance. The top 6 most critical channels (C4, O1, F8, O2, Fp1, P7) are highlighted with black circles. The prominence of Occipital (O1, O2) and Frontal (Fp1, F8) channels aligns with the pathological markers of alpha rhythm degradation and frontal slowing in Alzheimer's disease.}
\label{brain}
\end{figure}

The occlusion analysis identified specific spectral and spatial features driving the model’s decisions. Spectral perturbation revealed that Alpha (8–13 Hz) and Theta (4–8 Hz) are the most critical frequency bands (Fig. \ref{importance}(a)), with ROC AUC drops of 8.91\% and 7.85\%. Regarding electrode importance, C4 and O1 emerged as the most influential channels (Fig. \ref{importance}(b)), each causing a 7.45\% performance drop. Other high-ranking sensors included F8 (5.05\%), O2 (4.79\%), and Fp1 (4.79\%), indicating a concentration of predictive power in the occipital and frontal regions (Fig. \ref{brain}).

\section{Conclusion and Discussion}
This study establishes a high-performance computational framework for the detection of dementia using standard EEG. By integrating the representational power of the Large Brain Model (LaBraM) with a non-linear Random Forest classifier, we achieved a state-of-the-art ROC-AUC of 89.36\% and a Balanced Accuracy of 82.44\%. Adopting the Random Forest was critical, as it enabled the effective separation of complex, non-linear pathological signatures that standard linear projections could not resolve. Furthermore, the model attained a PR-AUC of 81.45\%, demonstrating that deep latent embeddings yield highly reliable diagnostic precision even in small, imbalanced datasets.

This study clarifies the hidden patterns captured by LaBraM. The model’s reliance on Alpha and Theta bands confirms it captures spectral slowing, a primary clinical hallmark of AD. In addition, the clinical alignment analysis demonstrates that higher LaBraM-predicted dementia probability correlates with lower MoCA scores, indicating that the model output reflects cognitive impairment rather than only diagnostic group separation. Furthermore, the significance of low-frequency activity supports the model’s sensitivity to pathological shifts. Our findings suggest that these shifts are driven by aperiodic background changes rather than periodic peaks, indicating that AD pathology manifests as global broadband neural slowing, a nuance often overlooked by traditional band-power analysis.

Topographically, the most influential regions—Occipital (O1, O2), Frontal (Fp1, F8), Central (C4), and Temporal (P7)—align with neural networks governing sensory-visual integration, executive function, and memory processing. The prominence of O1 and O2 directly supports the model's sensitivity to alpha rhythm degradation, while the importance of Fp1 and F8 reflects the frontal slowing characteristic of AD. By identifying these spatial-spectral biomarkers, this study provides a clinically interpretable framework for the diagnosis of AD.

A limitation of this study is the heterogeneity of the Control group, which includes individuals with Subjective Cognitive Decline (SCD). Although SCD can serve as a prodromal precursor to dementia, potentially introducing label noise, our model maintained high specificity (85.7\%). This resilience suggests that the framework is robust enough to distinguish objective neurophysiological markers of dementia from the subtler, subjective complaints associated with SCD. 

Another limitation is that the model's sensitivity (78.8\%) was lower than its specificity (85.7\%). This distinction is important to evaluate when considering the potential use as a screening tool, where sensitivity is particularly crucial for identifying at-risk individuals. Furthermore, although our data were multi-center, the cohort was recruited from a single geographic region. Future work should incorporate broader demographic diversity and cross-device hardware validation to confirm the generalizability of these foundation model embeddings across diverse ethnic populations and varying EEG recording systems. 

A further direction is to extend this framework to the detection of Mild Cognitive Impairment (MCI)\cite{jiang2025optimizing}. Investigating this prodromal phase is critical for defining the trajectory of neurodegeneration. We aim to analyze whether the latent topological features identified in this study can discriminate the transitional signal characteristics of MCI, thereby enabling intervention strategies before the onset of irreversible dementia. 

\bibliographystyle{IEEEtran}
\bibliography{reference} 

\end{document}